\pdfoutput=1

\documentclass[11pt]{article}

\usepackage[final]{emnlp2021}

\usepackage{times}
\usepackage{latexsym}
\usepackage{amsmath}
\usepackage{lipsum}
\usepackage[T1]{fontenc}

\usepackage[utf8]{inputenc}
\usepackage{microtype}
\usepackage{booktabs}
\usepackage{graphicx}
\usepackage{algorithm}
\usepackage{amsfonts}
\usepackage{algorithmic}
\usepackage{bbm}
\usepackage{comment}
\usepackage{multicol}  
\usepackage{multirow} 
\let\ACMmaketitle=\maketitle
\renewcommand{\maketitle}{\begingroup\let\footnote=\thanks \ACMmaketitle\endgroup}
\title{Contrastive Representation Learning for Exemplar-Guided Paraphrase Generation \footnote{The work described in this paper is substantially supported by a grant from the Research Grant Council of the Hong Kong Special Administrative Region, China (Project Code: 14200620).}}

\author{Haoran Yang\textsuperscript{1} , Wai Lam\textsuperscript{1} , Piji Li\textsuperscript{2}  \\
  \textsuperscript{1}The Chinese University of Hong Kong \\
  \textsuperscript{2}Tencent AI Lab \\ 
  \texttt{\{hryang,wlam\}@se.cuhk.edu.hk} \\
  \texttt{lipiji.pz@gmail.com} \\
  }

\begin{document}
\maketitle

\setlength{\abovedisplayskip}{0.6pt}
\setlength{\belowdisplayskip}{0.6pt}
\setlength{\abovedisplayshortskip}{0.6pt}
\setlength{\belowdisplayshortskip}{0.6pt}

\begin{abstract}

Exemplar-Guided Paraphrase Generation (EGPG) aims to generate a target sentence which conforms to the style  of the given exemplar while encapsulating the content information of the source sentence.
In this paper, we propose a new method with the goal of learning a better representation of the style and the content. This method is mainly motivated by the recent success of contrastive learning which has demonstrated its power in \textit{unsupervised} feature extraction tasks. The idea is to design two contrastive losses 
with respect to the content and the style 
by considering two problem characteristics during training. 
One characteristic is that the target sentence shares the same content with the source sentence, and the second characteristic is that the target sentence shares the same style with the exemplar. These two contrastive losses are incorporated into the general encoder-decoder paradigm. 
Experiments on two datasets, namely QQP-Pos and ParaNMT, demonstrate the effectiveness of our proposed constrastive losses. The code is available at \url{https://github.com/LHRYANG/CRL_EGPG}.

\end{abstract}

\section{Introduction}
Paraphrase generation~\citep{vaepara-2017,decpara-2019}, aiming to generate a sentence with the same semantic meaning of the source sentence, has achieved a great success in recent years. 
To obtain a paraphrase sentence with a particular style, Exemplar-Guided Paraphrase Generation (EGPG)~\citep{chen-control-2019} has attracted considerable attention. 
Different from other controllable text generation tasks whose constraints are taken from a finite set, e.g.,  binary sentiment or political slant~\citep{senti-2018,politics-2018}, multiple personas~\citep{persona-2019}, over which a classifier can be trained to guide the disentanglement process, the constraints of EGPG are exemplar sentences that can be arbitrarily provided, making it more challenging to learn a good representation for the style and the content. For example, as shown in Table~\ref{tab:exmple}, when using the content embedding of $X$ to retrieve a sentence with the most similar content in the target sentences list, we observe that the ordinary model, which is described in Section~\ref{abl}, can often match the sentence $Y^{\prime}$ whose content differs from $X$ instead of the correct target sentence $Y$. This reveals that the content encoder cannot encode the content information of a sentence appropriately which can result in inconsistent content of the generated sentence. 
The same problem also exists in the style encoder. 
\begin{table}[t]
\setlength{\abovecaptionskip}{4pt}
\centering
 \scalebox{0.7}{\begin{tabular}{ll}
\toprule
\vspace{0.2cm}
source ($X$)&\textit{what is the easiest way to get followers on quora ?}\\
exemplar ($Z$)&\textit{how do i avoid plagiarism in my article ?}\\
\midrule
target ($Y$)& \textit{how do i get more followers for my quora ?}\\
retrieved ($Y^{\prime}$)& \textit{what are the better ways to ask questions on quora ?}\\
\bottomrule[1pt]
\end{tabular}}
\caption{An example of EGPG}
\label{tab:exmple}
\vspace{-1em}
\end{table}

To learn a better content and style representations, we explore the 
incorporation of  \textit{contrastive learning} in EGPG to design an end-to-end encoder-decoder paradigm with multiple losses. Contrastive learning originates from  computer vision area~\citep{simclr-2020,supclr-2020} and now, it also shows its powerfulness  in natural language processing area. For instance,~\citet{iter2020pretraining} employ contrastive learning to improve the quality of discourse-level sentence representations.
In our proposed model, besides the basic encoder-decoder generation task, a content contrastive loss is designed to force the content encoder to distinguish features of the same content from features of different content. Similarly, a style contrastive loss is also employed to obtain a similar distinguishing effect for the style features. 
Experimental results on two benchmark datasets, namely QQP-Pos and ParaNMT, show that superior performance can be achieved with the help of the contrastive losses.

\section{Related Work}
\textbf{Paraphrase Generation} Researches on paraphrase generation has been for a long time. Traditional methods solve this problem mainly through statistical  machine   translation~\citep{stm-2004} or rule-based word substitution~\citep{smt-2010}. 
In the era of deep learning, approaches based on the encoder-decoder framework have emerged in large numbers~\citep{slstm-2016,chen2020distilling}. In addition to basic seq2seq model, ~\citet{li-etal-2018-delete} add a pair-wise discriminator to judge whether the input sentence and generated sentence are paraphrases of each other, with the help of reinforcement learning.   To generate diverse paraphrases, i.e., one to many mapping,~\citet{vaepara-2017} combine the power of RNN-based sequence-to-sequence model and the variational autoencoder. At decoding time, a noise sampled from the Gaussian distribution are appended to input to generate a diverse output.~\citet{qian-etal-2019-exploring}  propose a approach which use multiple generators to generate diverse paraphrases without sacrificing quality.

\noindent
\textbf{Exemplar-Guided Paraphrase Generation}
Making the generated paraphrases satisfy the style of an exemplar sentence is recently a hot research topic. EGPH is similar to other controlled text generation tasks whose constraints are sentiment~\citep{senti-2018,xu2021change}, 
gender~\citep{politics-2018}, topics~\citep{weige}. 
These tasks are highy related to Disentangled Representation Learning (DRL) which maps different aspects of the input data to independent low-dimensional spaces~\citep{cheng-etal-2020-improving}. 
~\citet{advctpara-2018} and ~\citet{syntax-2020} directly utilise the parse tree information of the exemplar as the style information without separating style from sentences. 
~\citet{chen-control-2019} propose a model which can directly extract style features from a modified target sentence. ~\citet{goyal-durrett-2020-neural} provide a way to generate paraphrase which is a component rearrangement of the original input through manipulating the parse tree.  

 
\noindent
\textbf{Contrastive Learning}
In the past few years, many unsupervised feature extraction algorithms have emerged, for instance, variational autoencoder~\citep{vae-2014,xu2020neural,gao2019discrete,gao2019generating}, generalised language models~\citep{gpt3-2019,bert-2019}. All the above methods obtain the feature of input by reconstructing the original input or predicting masked words and so on which do not take the relationships between the inputs into consideration. Therefore, contrastive learning, whose loss is designed to narrow down the distance between features of similar inputs and to enlarge the distance of dissimilar inputs, has been proposed and achieved a great success in both unsupervised~\citep{simclr-2020} and supervised~\citep{supclr-2020} image feature extraction. There are also some works trying to apply contrastive learning into natural language processing domain. For instance, ~\citet{iter2020pretraining} propose a pretraining method for sentence representation which employs contrastive learning to improve the quality of discourse-level representations. ~\citet{Giorgi2020DeCLUTRDC} utilise it to pretrain the transformer and btains state-of-the-art results on SentEval~\citep{conneau-kiela-2018-senteval}.
All of the above successes spur us to test whether contrastive learning is helpful on EGPG.

\section{Proposed Model}
\begin{figure}[t]
 \centering
  \includegraphics[width=0.8\linewidth]{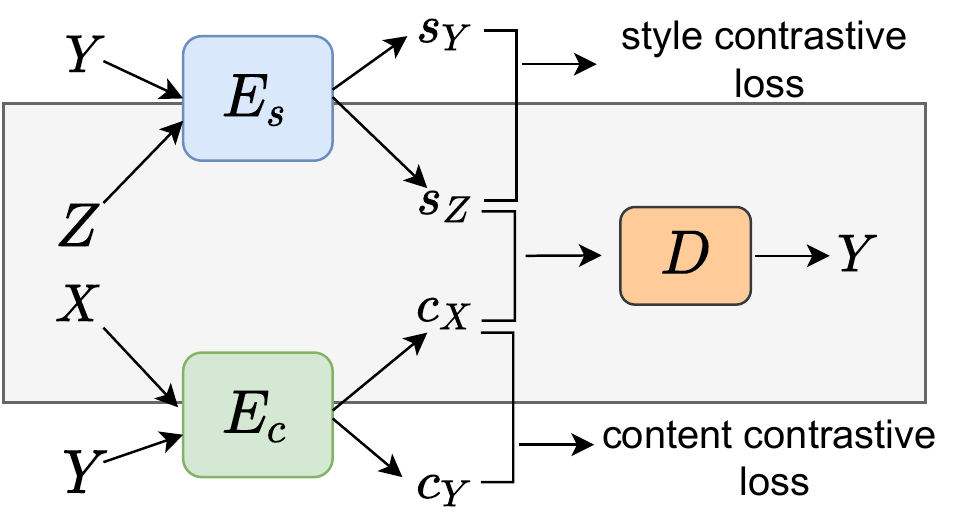}
  \caption{An Overview of Our Model
  }
  \label{fig:overview_of_our_model}
\end{figure}

Given a source sentence $X_i$ and an exemplar sentence $Z_i$, our goal is to generate a sentence $Y_i$ that has the same style (syntax) with $Z_i$ and retains the content (semantics) of $X_i$. 
As shown in Figure~\ref{fig:overview_of_our_model}, we design the encoders $E_s$ and $E_c$ for style and content respectively. The decoder $D$ generates the output.
Our model is trained by optimizing three losses simultaneously: (1) generation loss; (2) content contrastive loss; (3) style contrastive loss. 

\textbf{Generation Task}
For $X_i$ and $Z_i$, we firstly obtain their corresponding content features $c_{X_i}$ and style features $s_{Z_i}$: 
\begin{align}
    &c_{X_i} = E_c(X_i)\\
    &s_{Z_i} = E_s(Z_i)
\end{align}
Then $c_{X_i}$ and $s_{Z_i}$ are concatenated and inputted into the decoder as the initial hidden state to generate a sequence of probabilities over vocabulary. At the step $t$, the predicted probability $p_t$ of the $t$-th target word is obtained as follows
\begin{align}
    &p_t = softmax(Wh_t) \\
    &h_t = GRU(h_{t-1},e(y_{t-1}))
\end{align}
where $h_0$ is initialized as $\left[c_{X_i},s_{Z_i}\right]$ and $W$ is a parameter matrix. $y_{t-1}$ is the word in the previous step $t-1$ and $y_0$ is the special symbol [SOS] which represents the start of the sentence. $e(y_{t-1})$ is the embedding of the word $y_{t-1}$.

Negative log-likelihood loss (NLL) is employed as the basic optimization objective 
\begin{equation}
\mathcal{L}_i^{nll} = - \frac{1}{|Y_i|} \sum\limits_{t=1}^{|Y_i|} I(y_t)^T \log{p_{t}} 
\end{equation}
where $I(y_t)$ represents the one-hot encoding of the word $y_t$ in the vocabulary.

\textbf{Content Contrastive Learning (CCL)} Considering that $X_i$ and $Y_i$ share the same content, their content features should be close with each other in the content feature space.  Contrastive Learning which is designed to minimize the distance between positive pairs and maximize the distance between negative pairs can help model this relationship. 
Formally, during training, given a batch $\left\{(X_i,Y_i,Z_i)\right\}_{i=1}^n$ where $n$ is the batch size, we firstly obtain the corresponding content features of $X_i$ and $Y_i$, denoted by $\left\{(c_{X_i},c_{Y_i})\right\}_{i=1}^n$.
For $c_{X_i}$,  the positive pair is $(c_{X_i},c_{Y_i})$ and $c_{X_i}$ with the other remaining features in this batch form $2n-2$ negative pairs. For $c_{Y_i}$, the definition of positive/negative pairs is the same as $c_{X_i}$.  Then the contrastive loss  is employed, giving
\begin{equation}
    \mathcal{L}_{X_i}^{ccl} = -log\frac{exp(c_{X_i}\cdot c_{Y_i}/\tau)}{exp(\frac{c_{X_i}\cdot c_{Y_i}}{\tau})+\sum \limits_{j\neq i \atop T\in \left\{X,Y\right\}}exp(\frac{c_{X_i}\cdot c_{T_j}}{\tau})}
\end{equation}
\begin{equation}
     \mathcal{L}_{Y_i}^{ccl} = -log\frac{exp(c_{Y_i}\cdot c_{X_i}/\tau)}{exp(\frac{c_{Y_i}\cdot c_{X_i}}{\tau})+\sum \limits_{j\neq i \atop T\in \left\{X,Y\right\}}exp(\frac{c_{Y_i}\cdot c_{T_j}}{\tau})} 
\end{equation}
\begin{equation}
    \mathcal{L}^{ccl} = \sum_{i=1}^{n}(\mathcal{L}_{X_i}^{ccl}+\mathcal{L}_{Y_i}^{ccl})
\end{equation}
where · represents the dot product between two vectors and $\tau$ denotes a temperature parameter.

\textbf{Style Contrastive Learning (SCL)} aims to help $E_s$ learn a better style representation by considering that $Z_i$ and $Y_i$ share the same style. Similar to CCL, we firstly obtain the style features  $\left\{(s_{Z_i},s_{Y_i})\right\}_{i=1}^n$ and then apply the contrastive loss to these features
\begin{align}
 &\mathcal{L}_{Y_i}^{scl} = -log\frac{exp(s_{Y_i}\cdot s_{Z_i}/\tau)}{exp(\frac{s_{Y_i}\cdot s_{Z_i}}{\tau}) +\sum \limits_{j\neq i \atop T\in \left\{Z,Y\right\}}exp(\frac{s_{Y_i}\cdot s_{T_j}}{\tau})} \\ 
     &\mathcal{L}_{Z_i}^{scl} = -log\frac{exp(s_{Z_i}\cdot s_{Y_i}/\tau)}{exp(\frac{s_{Z_i}\cdot s_{Y_i}}{\tau}) + \sum \limits_{j\neq i \atop T\in \left\{Z,Y\right\}}exp(\frac{s_{Z_i}\cdot s_{T_j}}{\tau})} 
\end{align}
\begin{equation}
    \mathcal{L}^{scl} = \sum_{i=1}^{n}(\mathcal{L}_{Y_i}^{scl}+\mathcal{L}_{Z_i}^{scl})
\end{equation}
As a result, the total loss for a batch is as follows
\begin{equation}
    \mathcal{L} = \sum_{i=1}^n \mathcal{L}_i^{nll}+\lambda_1 \mathcal{L}^{ccl} + \lambda_2\mathcal{L}^{scl}
\end{equation}

\section{Experiments}

\subsection{Datasets}
We conduct experiments on two benchmark datasets, namely ParaNMT~\citep{chen-control-2019} and QQP-Pos~\citep{syntax-2020}. ParaNMT consists of about 500k training, 800 testing and 500 validation sentence pairs which are automatically generated through backtranslation of the original English sentences.
QQP-Pos consists of 130k training, 3k testing and 3k validation quora question pairs which are more formal than ParaNMT. The split size is the same as previous works to have a fair comparison. Since the exemplar sentences are not provided in both datasets, we adopt a method similar with~\citet{syntax-2020} to search an exemplar $Z_i$ for each source-target pair $(X_i,Y_i)$ based on the POS tag sequence  \footnote{We use NLTK for POS tagging.} similarity (refer to Appendix~\ref{sec:appendixA}). 
\vspace{-0.5em}
\subsection{Baselines \& Metrics}
We compare our model with (1) SCPN~\cite{advctpara-2018} which employs a parse generator to output the full linearized parse tree as the style by inputting a parse template; (2) SGCP~\citep{syntax-2020} which extracts the style information directly from the parse tree of the exemplar sentence; (3) CGEN~\citep{chen-control-2019}, an approach based on variational inference~\citep{vae-2014}. 

The evaluation metrics are  BLEU~\citep{BLEU}, METEOR~\citep{meteor} and ROUGE (R)~\citep{rouge}. We also conduct human evaluation to investigate the quality of the generated sentences. Moreover, we propose \textbf{Content Matching Accuracy} (CMA) to gauge the quality of the generated embeddings for the content. CMA will be introduced in Section~\ref{abl}.
\subsection{Implementation Details}
Each sentence is trimmed with a maximum length 15. The word embedding is initialized with 300-$d$ pretrained GloVe~\citep{glove}. We use a BERT-based~\citep{bert-2019} architecture for the style encoder $E_s$ and the dimension of style features is 768. For content encoder $E_c$, we use GRU~\citep{gru-2014} with hidden state size 512. During training, the teacher forcing technique is applied with the rate 1.0. 
The balancing parameters $\lambda_1$ and $\lambda_2$ are both set to 0.1. The temperature parameter $\tau$ is set to 0.5. 
 We train our model using Adam optimizer with the learning rate 1e-4 and the training epochs are set to 30 and 45 for ParaNMT and QQP-Pos respectively. 
 \subsection{Results}
As summarized in Table~\ref{tab:auto}, our model outperforms SCPN, SGCP and CGEN by a large margin on automatic evaluation metrics. We also conduct human evaluation to investigate the holistic quality of the generated sentences. For each dataset, we firstly choose two source sentences and then randomly select 25 exemplars for each source sentence to generate a total of 50 sentences. Table~\ref{tab:human} shows the results of human assessment. It can be seen that our model obtains a higher score than SGCP and CGEN, which is consistent with the automatic evaluation results. These results are expected because SCPN and SGCP use a parse tree as the style which is lack of the lexical information and very unstable. Moreover, CEGN is VAE-based which is intrinsically harder to train~\citep{bowman-etal-2016-generating}. 
\subsection{Ablation Study}
\label{abl}
 We conduct ablation study with three variants, namely ours without SCL (Ours-w/o-SCL), ours without CCL (Ours-w/o-CCL), ours without both SCL and CCL (Ours-w/o-both). We show that our model is better than the three variants to demonstrate the effectiveness of the contrastive losses. 

 \begin{table}[t]
 \setlength{\abovecaptionskip}{1pt}
    \large 
    \centering
    \resizebox{1.0\linewidth}{!}{\begin{tabular}{c|ccccc}
        \hline 
        \hline 
        \multicolumn{6}{c}{QQP-Pos} \\
        \hline
        \textbf{Model}& \textbf{BLEU}&\textbf{R-1}&\textbf{R-2}&\textbf{R-L}&\textbf{METEOR}\\
        \hline 
        SCPN&15.6&40.6&20.5&44.6&19.6\\
        SGCP&36.7&66.9&45.0&69.6&39.8\\
        CGEN&34.9&62.6&42.7&65.4&37.4\\
        \hline
        \hline 
        Ours&\textbf{45.8}&\textbf{71.0}&\textbf{52.8}&\textbf{73.3}&\textbf{45.8}\\
        \hline 
        Ours-w/o-CCL&43.1&70.0&50.6&72.3&43.5\\
        Ours-w/o-SCL&42.7&69.7&49.9&71.8&43.6\\
        Ours-w/o-both&40.8&68.4&48.4&70.8&41.6\\
        \hline
        \hline
        \multicolumn{6}{c}{ParaNMT} \\
        \hline 
         SCPN&6.4&30.3&11.2&34.6&14.6 \\
        SGCP&15.3&46.6&21.8&49.7&25.9\\
        CGEN&13.6&44.8&21.0&48.3&24.8\\
        \hline 
        Ours&\textbf{16.2}&50.6&\textbf{25.3}&52.1&\textbf{28.4}\\
        \hline 
        Ours-w/o-CCL&15.3&\textbf{50.8}&25.2&\textbf{52.4}&28.0\\
        Ours-w/o-SCL&15.2&50.2&24.4&51.5&28.0\\
        Ours-w/o-both&15.3&50.2&24.9&51.6&27.7 \\
        \hline 
        \hline 
        
    \end{tabular}
    }
    \caption{Automatic Evaluation Results.}
    \label{tab:auto}
\end{table}

\begin{table}[t]
\setlength{\abovecaptionskip}{1pt}
    \footnotesize
    \centering
    \resizebox{1.0\linewidth}{!}{\begin{tabular}{ccccccc}
    \hline 
    Model&Ours&Ours-w/o-CCL&Ours-w/o-SCL&Ours-w/o-both \\ 
    \hline 
    \multicolumn{5}{c}{QQP-Pos}\\
    \hline 
    ED-E&2.49&2.42&2.56&2.57\\ 
    ED-R&2.64&2.65&2.78&2.82\\
    \hline 
    \multicolumn{5}{c}{ParaNMT}\\
    \hline 
    ED-E&4.36&4.47&4.49&4.49\\ 
    ED-R&4.22&4.24&4.28&4.25\\
    \hline 
    \end{tabular}}
    \caption{Style Evaluation}
    \label{tab:style_eva}
    \vspace{-1em}
\end{table}

\begin{table*}[t]
    \setlength{\abovecaptionskip}{1pt}
    \centering
     \resizebox{\linewidth}{!}{\begin{tabular}{ll}
        \hline  
        \multicolumn{2}{c}{\textbf{SOURCE:} how do i develop good project management skills ?}\\
        \hline
        \textbf{EXEMPLAR} & \textbf{GENERATION} \\
        \hline 
         which is the best laptop model to buy within 30k ? & which is the best way to develop project management skills ? \\
         how many cups of coffee should i consume in a day ?& from what skills can i start in a project management ? \\
         which subjects are important to become a chartered accountant ?& what skills are necessary to develop a project management ?\\
         \hline 
    \end{tabular}}
    \caption{An Example Generated by Our Model}
    \label{tab:exm}
    \vspace{-4pt}
\end{table*}
\begin{table*}[h]
\setlength{\abovecaptionskip}{1pt}
    \footnotesize
    \centering
    \begin{tabular}{ccccccc}
    \hline 
    Model&Ours&SGCP&CGEN&Ours-w/o-CCL&Ours-w/o-SCL&Ours-w/o-both \\ 
    \hline 
    QQP-Pos&\textbf{3.71}&3.12&2.97&3.59&3.45&3.33\\ 
    ParaNMT&\textbf{2.53}&1.9&2.05&2.51&\textbf{2.53}& 2.38\\
    \hline 
    \end{tabular}
    \caption{Human evaluation results. Each sentence is given a score ranging from one to five to assess the holistic quality.  We report the average value of two annotators. Higher score is better. The Spearman's correlation coefficients of these two annotators are 0.707 for QQP-Pos and 0.37 for ParaNMT.}
    \label{tab:human}
    \vspace{-6pt}
\end{table*}
\begin{table}[h]
\setlength{\abovecaptionskip}{4pt}
\centering
 \scalebox{0.7}{\begin{tabular}{ll}
\toprule
$X$&can you view a private facebook profile? \\
$Z$&how would you learn a new programming language?\\
\midrule
 $Y$&how do you view a private facebook profile?\\
 $Y^{\prime}$& can you see who visited your instagram last? \\
\midrule 
Ours&how do you see a private facebook profile?\\
Ours-w/o-SCL& how can you view a private facebook profile? \\ 
Ours-w/o-both& how can you view a private instagram profile?\\
\bottomrule[1pt]
\end{tabular}}
\caption{A failed example without CCL}
\label{tab:exm_ccl}
\end{table} 
As presented in Table~\ref{tab:auto}, we can see that Ours can achieve better results than the three variants on all automatic metrics for QQP-Pos. Particularly, Ours, Ours-w/o-CCL and Ours-w/o-SCL outperform Ours-w/o-both a lot, demonstrating the usefulness of the contrastive losses.    For ParaNMT, Ours obtains the highest score on BLEU, ROUGE-2, MENTOR while it lags behind Ours-w/o-CCL on ROUGE-1 and ROUGE-L. This phenomena may be caused by the poor quality of the dataset. 
The human evaluation results are also listed in Table~\ref{tab:human}. Our model can generally generate fluent sentences on QQP-Pos. But the overall quality of sentences generated by all models on ParaNMT is unsatisfactory which shows that a high-quality dataset is necessary for training a good model.  


We also provide the style evaluation of \textbf{ED-E} (edit distance between the POS tag sequence of the generated paraphrase and the exemplar) and \textbf{ED-R} (edit distance between the POS tag sequence of the generated paraphrase and the ground truth reference) in Table~\ref{tab:style_eva}. We can see that models with style contrastive losses have smaller edit distance. 

\begin{figure}[t]
\setlength{\abovecaptionskip}{1pt}
    \centering
    \includegraphics[width=0.65\linewidth]{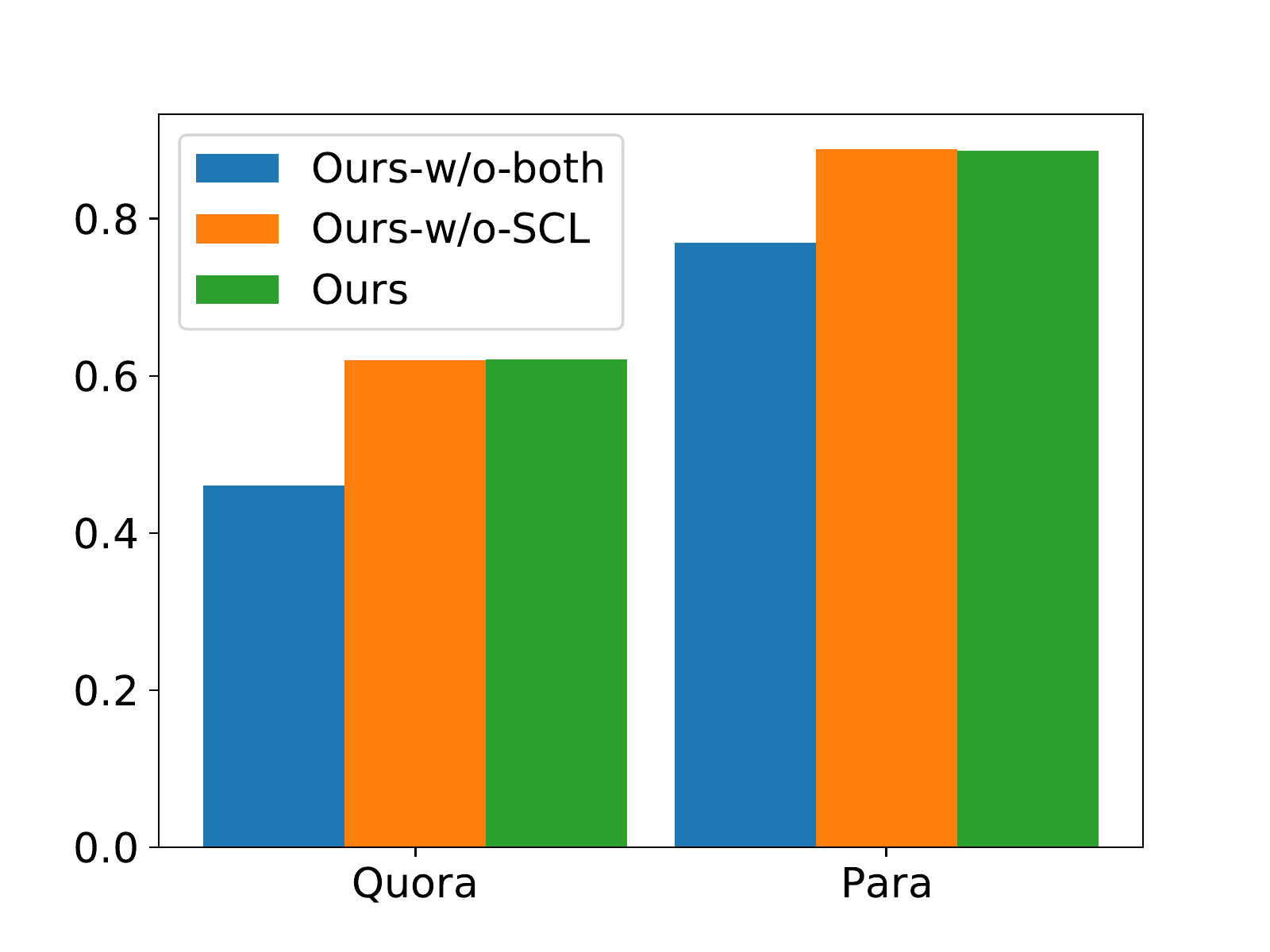}
    \caption{Content Matching Accuracy}
    \label{fig:vis}
    \vspace{-6mm}
\end{figure}

To directly assess the quality of the generated embeddings for the content we propose \textbf{Content Matching Accuracy}. To calculate CMA, firstly we input all the source and target sentences into $E_c$ to get the content representations $\textbf{A}, \textbf{B} \in \textbf{R}^{m\times k}$, where $m$ is the size of the test dataset and $k$ is the dimension of content feature vectors. Then, we calculate the similarity matrix $\textbf{S} = \textbf{A}\textbf{B}^T$. In ideal situation, each diagonal element $\textbf{S}_{[i,i]}$ should be the largest value in row $i$ since the content embedding $\textbf{A}_i$ of the $i_{th}$ source sentence should have the  greatest similarity with  the content embedding $\textbf{B}_i$ of its corresponding target sentence.
Therefore, the content matching accuracy (CMA) is defined as:

\begin{equation}
    CMA = \frac{\sum_{i=1}^{m}\mathbbm{1}(argmax(\textbf{S}_i) = i)}{m}
\end{equation}
where $\mathbbm{1}(s)$ equals 1 if $s$ is true, otherwise 0. 
The results are illustrated in Figure~\ref{fig:vis}. We notice that models with CCL can achieve higher accuracy than Ours-w/o-both. It signifies that the content encoder $E_c$ is improved with the help of CCL. We provide a failed example in Table~\ref{tab:exm_ccl}. $Y^{\prime}$ is the sentence retrieved given $X$ under the model Ours-w/o-both. $Y$ is the target sentence and it is also the sentence retrieved given $X$ under the model Ours or Ours-w/o-SCL. We can see that the content generated by Ours-w/o-both is incorrect (changing facebook to instagram) and instagram exists in $Y^{\prime}$. This illustrates that the poor-quality content embedding of $X$ can cause the incorrect content of the generated sentence. In general, matching accuray can also be calculated for the style. However, the exemplar selection process has a high probability of dropping the sentence with the most similar style of $Y$. Therefore, style matching accuracy is not provided here. Instead, we list some retrieved sentences based on the style embedding in Appendix~\ref{sec:appendixB}.    

\subsection{Case Study}
Some examples generated by our model are shown in Table~\ref{tab:exm}. It can be observed that our model can generally generate high-quality sentences which have similar style with the exemplar and retain the semantic meaning of the source sentence. Moreover, our model does not directly copy the style words from the exemplar, but instead adopts the overall structure of the exemplar to generate sentences, for example, the second one. More examples are provided in Appendix~\ref{sec:appendixC}.

\subsection{Explanations}
We attempt to provide some possible explanations about why the model with these two contrastive losses can achieve better performance. The first reason is that adding additional losses on the output of encoders can alleviate gradient vanishing which is a serious issue when training the encoder-decoder model. The second reason is that the overfitting issue may be prevented, since the contrastive losses restrict the free adjustment of parameters in the model by forcing the encoder and decoder to focus on their own tasks, i.e., feature extraction and sentence generation. 

\section{Conclusion}
We introduce the content contrastive loss and the style contrastive loss into EGPG to design a multi-losses scheme without requiring additional labeled data. This scheme can obtain better results compared with the baseline and ablative models, which demonstrates the effectiveness of contrastive learning for learning better representations. Moreover, the proposed framework is general and may  benefit other similar NLP tasks.

\bibliography{anthology,custom}
\bibliographystyle{acl_natbib}

\newpage 
\clearpage
\appendix

\section{Exemplar Searching Algorithm}
\label{sec:appendixA}
\begin{algorithm}[h]
    \small
    \caption{Searching Exemplar Sentences}
    \begin{algorithmic}[1]
        \REQUIRE dataset $\mathbb{D}=(\mathbb{D}_X,\mathbb{D}_Y)$ 
        \FOR{$Y$ in $\mathbb{D}_Y$}
        \STATE find the sentence set $\mathbb{C}_1 \subseteq \mathbb{D}_X$ that each $C \in \mathbb{C}_1$ satisfies $\left|len(C)-len(Y)\right| \leq 2$
        \STATE find the sentence set $\mathbb{C}_2\subseteq \mathbb{C}_1$ that for each $C \in \mathbb{C}_2$, the number of shared words between $C$ and $Y$, denoted by $c$, satisfies $c+2\leq len(Y)$  
        \STATE find the exemplar $Z\in \mathbb{C}_2$ which has the smallest POS tag sequence  editdistance  with $Y$
        \ENDFOR
    \end{algorithmic}
    \label{alg:A}
\end{algorithm}
The detailed steps for exemplar sentence searching are described in Algorithm~\ref{alg:A}. 
Step 2 is done to accelerate the searching procedure since sentences with similar style tend to have similar token lengths.
Step 3 guarantees that the content information of $Y$ and the selected $Z$ does not overlap much.

\section{Style Embedding Quality}
\label{sec:appendixB}
We list some sentences retrieved by ablative models given the style embedding of a sentence $S$ in Table~\ref{tab:style_embd}. For each model, we obtain the top-5 sentences which are most similar to $S$. We can see that the sentences retrieved by Ours and Ours-w/o-CCL are more similar to $S$ in style dimension than Ours-w/o-both on the whole. For example, in the second case, the fifth sentence of Ours-w/o-both lacks the adverbial modifier compared with $S_2$.
\begin{table}[h]
\setlength{\abovecaptionskip}{4pt}
\centering
 \scalebox{0.55}{\begin{tabular}{ll}
\toprule
$S_1$&what are newton 's laws of motion? \\
\midrule
 \multirow{5}{*}{Ours}&what are the after effects of masturbation ?\\
 &what are the health benefits of coffee ? \\
 &what are the safety precautions on handling shotguns ?\\
 &what are some interesting facts about bengaluru ?\\
 &what are some unknown facts about football ?\\ 
\midrule 
\multirow{5}{*}{Ours-w/o-CCL}&what are the health benefits of coffee ?\\
& what are the good things about pakistan ? \\ 
& what are some interesting facts about bengaluru ?\\
&what are some unknown facts about football ?\\
&what are considered abiotic factors of grasslands ?\\

\midrule 
\multirow{5}{*}{Ours-w/o-both}&what were nelson mandela 's greatest accomplishments ?\\
& what are craig good 's qualifications to talk about nutrition ? \\ 
&what are reasons of china 's success ?\\
&what are president obama 's greatest accomplishments and failures ?\\
&what is newton 's third low of motion with examples ?\\
\midrule
\midrule 
$S_2$&how do i impress a girl on chat ? \\
\midrule 
\multirow{5}{*}{Ours}&how do i become an engineer in robotics ?\\
& how do i get a job in europe countries ? \\ 
&how do i get the crown on musical ly ?\\
&how do i make a website responsive without bootstrap ?\\
&how do i find the best seo company in dellhi ncr ?\\

\midrule 
\multirow{5}{*}{Ours-w/o-CCL}&how do i become an engineer in robotics ?\\
& how do i get a job in europe countries ? \\ 
& how do i leave a girl without hurting her feelings ?\\
& how do i get the crown on musical ly ?\\
& how do i root a galaxy s550 at t ?\\

\midrule 
\multirow{5}{*}{Ours-w/o-both}&how do i become an engineer in robotics ?\\
& how do i get a job in europe countries ? \\ 
&how do i buy a suit online ?\\
&how do i get the crown on musical ly ?\\
&how can i help a friend get off drugs ?\\

\bottomrule[1pt]
\end{tabular}}
\caption{Sentences Retrieved by Style Embedding}
\label{tab:style_embd}
\vspace{-1.5em}
\end{table} 

\section{Multiple Paraphrase Sentences Generation}
\label{sec:appendixC}
\begin{table}[h]
\setlength{\abovecaptionskip}{4pt}
\centering
 \scalebox{0.6}{\begin{tabular}{cc}
\toprule
$X_1$&which is the best anime to watch ?\\
\midrule 
$Z_{11}$&can you jailbreak an ios 8 3 ?\\
$Y_{11}$&can you recommend the best anime ?\\
\midrule 
$Z_{12}$&which are the best mba colleges in gwalior ?\\
$Y_{12}$&what are the best anime films of all time ?\\
\midrule
$Z_{13}$&how can i earn from online ?\\
$Y_{13}$&what anime should i watch now ?\\
\midrule
\midrule
$X_2$&what type of music do you listen ?\\
\midrule 
$Z_{21}$&is equatorial guinea really rich ?\\
$Y_{21}$&which music is really good ?\\
\midrule 
$Z_{22}$&what tv series have you watched and why did you like them ? explain\\
$Y_{22}$&what type of music do you like ? and how do you recommend ?\\
\midrule
$Z_{23}$&what is the best way to reduce weight ?\\
$Y_{23}$&what is the best music to listen to ?\\
\bottomrule[1pt]
\end{tabular}}
\caption{Generate different sentences given different exemplars.}
\label{tab:multiple_generation}
\vspace{-1.5em}
\end{table} 

\end{document}